\documentclass[runningheads]{llncs}
\usepackage{graphicx}
\usepackage{tikz}
\usepackage{comment}
\usepackage{amsmath,amssymb} % define this before the line numbering.
\usepackage{color}

\usepackage[accsupp]{axessibility}  % Improves PDF readability for those with disabilities.
\usepackage{subcaption}

\begin{document}
% \linenumbers
\pagestyle{plain}
\mainmatter
% Insert your submission number here

\title{1st Place Solutions for the UVO Challenge 2022} % Replace with your title

%\begin{comment}
% \titlerunning{ECCV-22 submission ID } 
% \authorrunning{ECCV-22 submission ID } 
\author{Jiajun Zhang{$^{1\dag}$},~ Boyu Chen{$^{2\dag}$},~ Zhilong Ji{$^{2*}$},~ Jinfeng Bai{$^{2}$},~ Zonghai Hu{$^{1}$}}
% \author[author2]{Boyu Chen}  
% \author[author2]{Zhilong Ji}
% \author[author2]{Jinfeng Bai}
% \author[author1]{Zonghai Hu}
\institute{{$^{1}$}Beijing University of Posts and Telecommunications\\ {$^{2}$}Tomorrow Advancing Life \\
\{jiajun.zhang, zhhu\}@bupt.edu.cn, \{chenboyu, jizhilong, baijinfeng1\}@tal.com}
% \author{Jiajun Zhang{$^{1\dag}$},~ Boyu Chen{$^{2\dag}$},~ Zhilong Ji{$^{2*}$},~ Jinfeng Bai{$^{2}$},~ Zonghai Hu{$^{1}$}\\\vspace{-8pt}{\small~}\\
% {$^{1}$}Beijing University of Posts and Telecommunications\\ {$^{2}$}Tomorrow Advancing Life\\
% {\small{{$^{\dag}$}Contribute equally~~{$^{*}$}Corresponding to: \tt{liuxiaofengcmu@gmail.com}}}
% }
%\end{comment}
\def\thefootnote{$\dag$}\footnotetext{Contribute equally}
%******************

\begin{comment}
\titlerunning{Abbreviated paper title}
\author{First Author\inst{1}\orcidID{0000-1111-2222-3333} \and
Second Author\inst{2,3}\orcidID{1111-2222-3333-4444} \and
Third Author\inst{3}\orcidID{2222--3333-4444-5555}}
%
\authorrunning{F. Author et al.}
\institute{Princeton University, Princeton NJ 08544, USA \and
Springer Heidelberg, Tiergartenstr. 17, 69121 Heidelberg, Germany
\email{lncs@springer.com}\\
\url{http://www.springer.com/gp/computer-science/lncs} \and
ABC Institute, Rupert-Karls-University Heidelberg, Heidelberg, Germany\\
\email{\{abc,lncs\}@uni-heidelberg.de}}
\end{comment}
%******************
\maketitle

\begin{abstract}
This paper describes the approach we have taken in the challenge. We still adopted the two-stage scheme same as the last champion, that is, detection first and segmentation followed. We trained more powerful detector and segmentor separately. Besides, we also  perform pseudo-label training on the test set, based on student-teacher framework and end-to-end transformer based object detection.
The method ranks first on
the 2nd Unidentified Video Objects (UVO) challenge, achieving AR@100 of 46.8, 64.7 and 32.2 in the limited data frame track, unlimited data frame track and video track respectively.
\end{abstract}

\section{Introduction}
Common instance segmentation algorithms are trained on specific datasets to learn a fixed number of categories for specific scenarios. UVO \cite{uvo} discusses a more realistic scenarios, Open-World instance segmentation. In this scenario, we expect that the algorithm is capable of detecting or segmenting novel objects and have the ability of incremental learning.

Compared with the end-to-end instance segmentation framework, two-stage framework often achieves better performance. We have tried these two kinds of methods for comparison in our initial experiments, and even the end-to-end SOTA method Cascade Mask-RCNN \cite{maskrcnn} with backbone ViTDet \cite{vitdet} still has a significant gap with the two-stage method. Therefore, we adopted the two-stage scheme same as the last year's winner. As mentioned in \cite{20211st}, stwo-stage architecture enable us to train the detection network and segmentation network separately on different datasets and use more complex models. Obviously, there is a trade-off between accuracy and complexity.

To further improve the detection performance on test set and exhaustively detecting the unseen object/class. We define this problem as Semi-Supervised Object Detection (SSOD). Different from previous work, Soft-teacher \cite{softteacher} proposed an end-to-end training framework for semi-supervised object detection, student-teacher framework simultaneously improves the detector and pseudo labels by leveraging a student model for detection training, and a teacher model which is continuously updated by the student model through the exponential moving average strategy for online pseudo-labeling.

We will detail the detection, segmentation model and semi-supervised object detection method we used in the next chapter. 

\section{Method}

\subsection{Detection}
DETR \cite{detr} proposed a Transformer-based end-to-end object detector without using hand-designed components like anchor design and NMS, and achieves comparable performance with Faster-RCNN \cite{fasterrcnn}. Many following works continue to improve the DETR-like model, and finally make DETR-like model the current new SOTA for object detection. For example, 
Deformable DETR \cite{deformabledetr} predicts 2D anchor points and designs a deformable attention module that only attends to certain sampling points around a reference point; DAB-DETR \cite{dabdetr} further extends 2D anchor points to 4D anchor box coordinates to represent queries and dynamically update boxes in each decoder layer; DN-DETR \cite{dndetr} introduces a denoising training method to speed up DETR training. It feeds noise-added ground-truth labels and boxes into the decoder and trains the model to reconstruct the original ones. 

DINO \cite{dino} collects the above improvements, and introduce a contrastive way for denoising training, a mixed query selection method for anchor initialization, and a look forward twice scheme for box prediction. DINO achieves the best result of 63.2AP on COCO val2017, which is the current SOTA and show us its powerful performance. Hence, We adopt DINO as our detector with bakcbone SwinL \cite{swin}.  
% With Object365 \cite{objects365} pretrain and SwinL  backbone, DINO achieves the current best results of 63.2AP on COCO val2017. 
\subsection{Segmentation}

ViT \cite{vit} also performed well in the segmentation track, for example, SETR \cite{setr}, Segmenter \cite{segmenter}. Masked-attention Mask Transformer (Mask2Former) \cite{mask2former}, is a new architecture capable of addressing any image segmentation task, and outperforms of SOTA specialized architectures on all considered tasks and datasets. Its key components include masked attention, which extracts localized features by constraining cross-attention within predicted mask regions. It also use multi-scale high-resolution features which help the model to segment small objects. 

ViT-Adapter \cite{vita} design a powerful adapater for plain ViT. It consists of: 1) a spatial prior module to capture spatial features from the input image; 2) a spatial feature injector to inject spatial priors into the ViT; 3) a multi-scale feature extractor to extract hierachical features fromt the ViT. 

Therefore, We adopt Mask2Former \cite{mask2former} as our segmentor with backbone ViT-Adapter-L \cite{vita}. 
\subsection{Semi-Supervised Object Detection}

Prevalent SSOD paradigm is to use a multi-stage self-training pipeline: 1) train model on labeled data; 2) generate pseudo labels on unlabeled data; 3) retrain model on both labeled and pseudo-labeled data; 4) repeat this process if needed. Soft Teacher \cite{softteacher} proposed an end-to-end semi-supervised approach, in contrast to previous complex multi-stage method. The end-to-end training gradually improves pseudo label qualities during the curriculum, and the more and more accurate pseudo labels in turn benefit object detection training. We transplant the DINO model to the soft-teacher training framework, and finetune our detection model on UVO test dataset as unlabeled data, and UVO train/val dataset labeled.

\section{Dataset}
\subsection{Frame Track limited}
Only ImageNet \cite{imagenet} annotations and COCO17 \cite{cocodataset} annotations (bounding box, mask, category) are allowed in this track. 
\subsection{Frame Track unlimited} 
\textbf{Detection}: ImageNet is used to pretrain backbone network. We pre-train our detector on Object365 \cite{objects365} and COCO \cite{cocodataset} datasets. Then, detector is finetuned on UVO frameset trainval. Finally, the model will be trained with the pseudo-label of test set under soft-teacher framework to improve the performance on test set. Object365 is a large-scale detection dataset with over $1.7M$ annotated images for training, the experiments also proved that the pre-training of object365 resulted in a significant performance improvement. One of our observations is that using COCO pretrain did not lead to performance boosting when our model is trained enough Object365.

\noindent\textbf{Segmentation}: We cropped all instances which have segmentation annotation from OpenImage \cite{openimage} and COCO \cite{cocodataset} dataset. Our segmentation task thus becomes a fore/background semantic segmentation of a single object. We pretrain our model on OpenImage and COCO instances and finetune on UVO frameset trainval.

\subsection{Video Track}
Our experiments proved that video dense set contains a large number of duplicate samples which may lead the model overfitting on training set, results in the performance degradation on test set. Therefore, we use the same data setting as Frame Track Unlimited. It should be noted that, the amount of training, validation and test set has doubled, compared with the last challenge.

\section{Implementation Details}
\subsection{Detection}
DINO \cite{dino}   is composed of backbone,  transformer encoder, transformer decoder and prediction heads. To achieve SOTA results, we use the SwinL \cite{swin} pretrained on ImageNet as our backbone.  We use the same random crop and scale augmentation during training following DINO \cite{dino}.  For DINO-5scale, we extract features from stages 1, 2, 3 and 4 of the backbone and an extra feature by down-sampling the ouput of stage 4.  We first pre-train DINO on COCO \cite{cocodataset} with 36-epoch setting (also called 3x) and pre-train on Object365 \cite{objects365}  with 12-epoch setting (1x), then finetune it on UVO. DINO is pretrained in class-agnostic way with bakcbone learning rate $1e-4$, other hyper-parameters are same as DINO. 

\subsection{Segmentation}
We trained our Mask2former-ViT-Adapter with MMSegmentation \cite{mmseg}.  The patch size of ViT is fixed to 16. The feature interaction times of adapter is set to $N=4$. Deformable attention is adopted in two feature interaction operators, in which the number of sampling points is fixed to 4. In last interaction, three multi-scale feature extractors are stacked together. We employ an AdamW [55] optimizer with an initial learning rate of $2\times10^{-5}$, and weight decay of 0.05. Random horizontal flipping, random cropping as well as random color jittering are used as data augmentation. We use a crop size of $512 \times 512$, a batch size of 4 and train models for 320k and 80k iterations for OpenImage COCO pre-train and UVO finetune respectively. 

\subsection{Semi-Supervised Object Detection}
\begin{figure}
\centering
\includegraphics[width=0.8\linewidth]{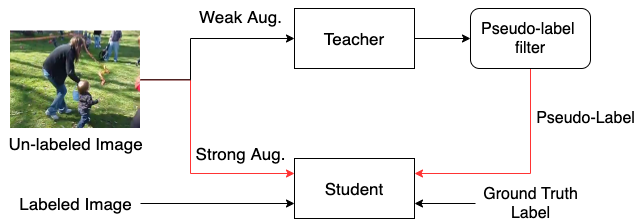}
\caption{Teacher-Student SSOD framework}
\label{fig:semi}
\end{figure}

Let us introduce training pipeline first. As shown in Fig.\ref{fig:semi}, for un-labeled image, two views of the image are generated by a strong and a weak augmentation. The weakly augmented view is forwarded through the teacher to produce the detection prediction. Then the filter generates the pseudo-labels, which are used to supervise the learning of the student on the strong augmentation.

There are two ways to filter out low-quality candidate boxes: confidence threshold and topK. We have done experiments on the two methods, and the results show that topK is better and we set $K=15$. Follwing \cite{omni-detr}, for strong augmentation, we apply horizontal flipping, random resizing, random size cropping, color jittering, grayscale, gaussian blur, and cutout patches; For weak augmentation, only random horizontal flipping is used. Even though we trained on 32G V100, we still restricted by GPU memory. The maximum size of image height and width is set to 1000, and set batch size of labeled data to 1. We believe that performance can be further improved if computing resources are not restricted. Finally, We evaluate our model on teacher model. 

\subsection{Video tracker}
We tried DeepSort \cite{deepsort} to track the detection bounding boxes, but the effect was not satisfactory. So we use last year winner's method, more details and discussion in the next section.

\section{Experiment Results}
\subsection{Frame Track Limited}
UVO dataset contains many non-coco objects, our semi-supervised method is dedicated to learning novel object.  After our detector is pretrained well on COCO, we use different part of UVO dataset as unlabeled data and test on validation dataset. In Tab.\ref{table:smi}, SSOD can improve a 62.4 AR@100 by + 3.9 AR@100, reaching 66.3 AR@100, by leveraging unlabeled data. Besides, we find that using more unlabeled data does not bring extra benefits, that may be caused by the big gap inside the UVO dataset.
\setlength{\tabcolsep}{3pt}
\begin{table}
\begin{center}
\caption{Different Part of dataset used for SSOD and eval on UVO validation dataset}
\label{table:smi}
\begin{tabular}{ll}
\hline\noalign{\smallskip}
Unlabeled dataset & AR@100\\
\noalign{\smallskip}
\hline
\noalign{\smallskip}
Without Semi-finetune  & 62.4 \\
Val &  \textbf{66.3} \\
Train + Val & 66.1 \\
\hline
\end{tabular}
\end{center}
\end{table}
\vspace{-1cm}
\subsection{Frame Track Unlimited}

\begin{figure*}[t!]
	\begin{subfigure}[t]{0.40\textwidth}
		\includegraphics[height=1.6in]{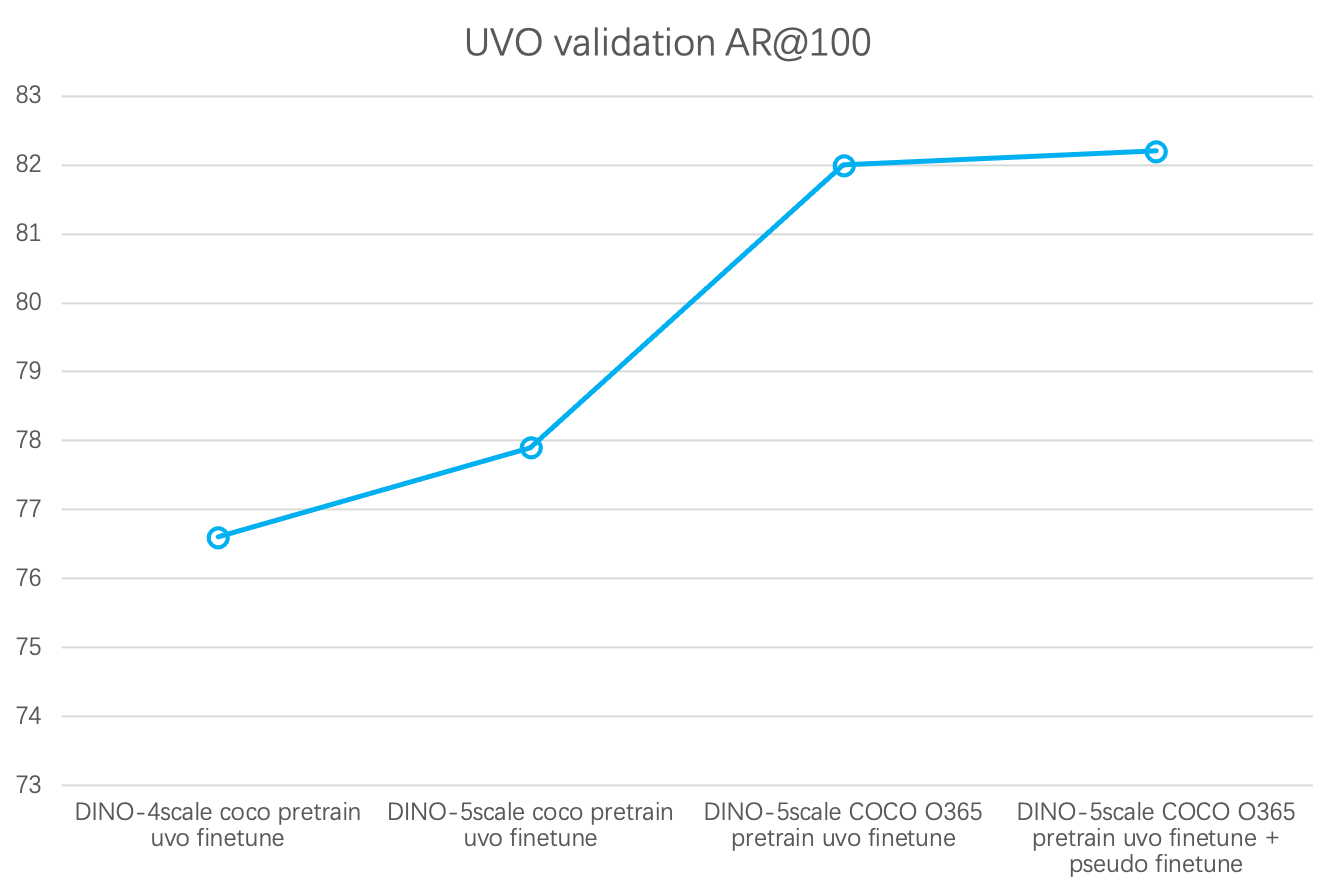}
		\label{fig:sega}
	\end{subfigure}
	\hspace{10mm}
	\begin{subfigure}[t]{0.40\textwidth}
		\includegraphics[height=1.6in]{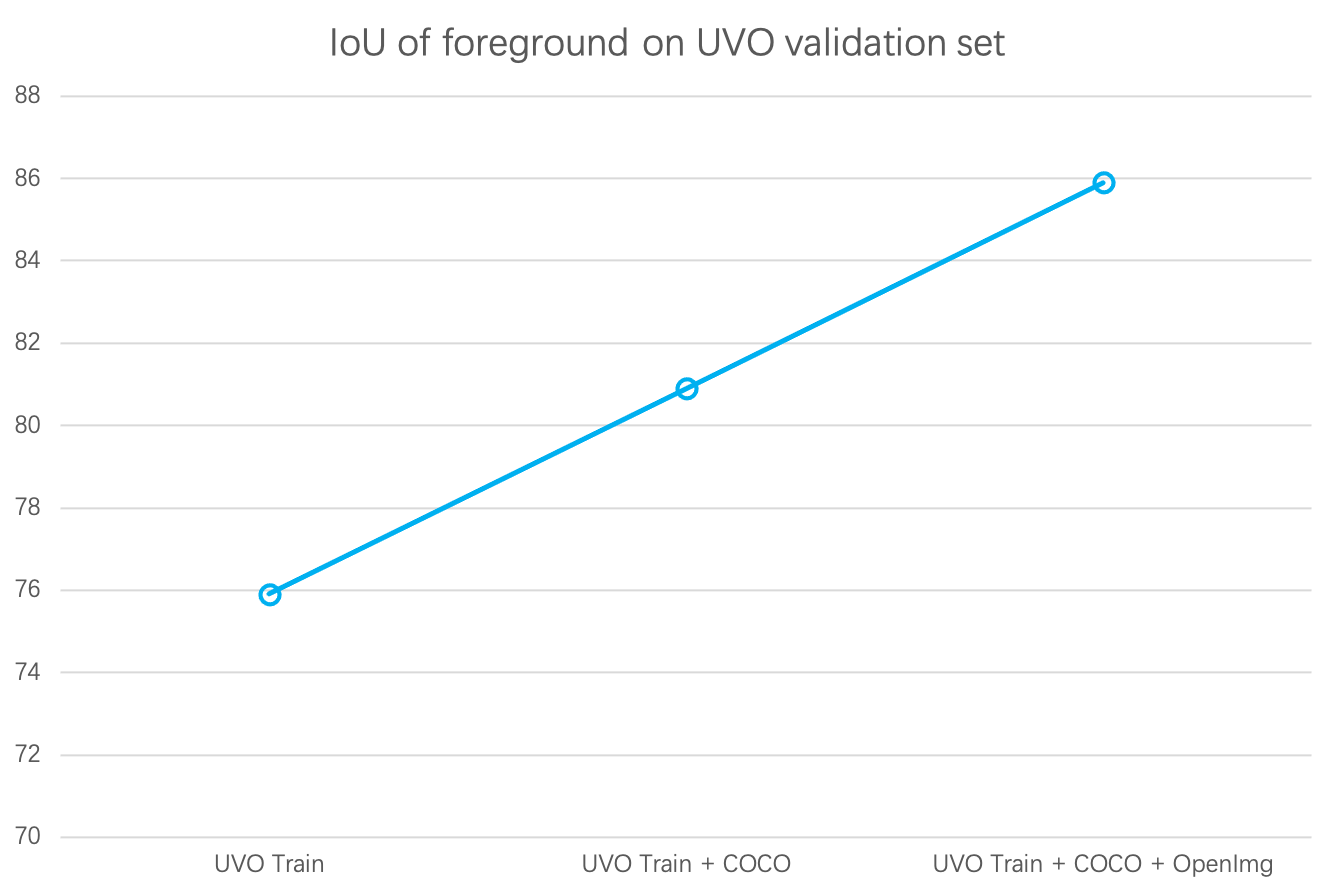}
		\label{fig:segb}
	\end{subfigure}
	\caption{UVO validation Bounding box AR and foreground IoU}
	\label{fig:seg}
\end{figure*}

Fig.\ref{fig:seg} demonstates the improvements we have made during training. DINO-4scale pretrained on COCO and fine-tune on UVO trainset is our baseline which achieved AR@100 of 76.6. By using 5scale and training with more epochs (3x), the performance could be improved by 1.3 points.  Object365 pretrain could boost the performance heavily by 4.1 points and achieved AR@100 of 82.0. With the final semi-supervised object detection strategy finetune on pseudo-label,  we slightly improved by 0.2 point and achieved AR@100 of 82.2 on UVO validation set. Compared with the obvious improvement in Tab.\ref{table:smi}, pseudo-label finetune did not improved that much, because the pre-trained model is already powerful.
\\

UVO annotated instance masks in a class-agnostic way, based on our experiment setup, our segmentation task aims to segment fore/background of a single object. As shown in Fig.\ref{fig:seg}, by increasing the amount of pretrained data, our segmentation performance is significantly improved.

\subsection{Video Track}
We meant to use a different tracker from last year's winner \cite{20211st}, but the results were not good. We tried the DeepSort \cite{deepsort} scheme as our tracker to track the detection bounding boxes. DeepSort is an online and real-time method based on tracking by detection. It performs Kalman filtering in
image space and frame-by-frame data association using the
Hungarian method with an association metric that measures
bounding box distances. The distance between each bounding box is measure by CNN features extracted from well-trained Re-identification model. We trained a simple classification model of Resnet50 \cite{r50} according to the label annotation from UVO. After adjust hyperparameters for several times, e.g. NMS thresh, confidence thresh and metric distance, the highest AR@100 we got is $20.8$, which is far from 32.2 by Du et al.\cite{20211st} tracking solution.

Hence, the final results we submitted, is still used last year's winner tracking scheme. Althogh DeepSort achieves a good performance in terms of tracking precision and accuracy, it returns a relatively high number of identity switches. This is, because the employed association metric is not accurate enough. UVO dataset annotate 80 coco categories, and for non-coco categories, they uniformly annotate as 1 category. For non-coco target, the features extracted from Reid model are close, which leads to unsatisfactory tracking results. 
\subsection{Visualization}
\begin{figure}
\centering
\includegraphics[width=1.0\linewidth]{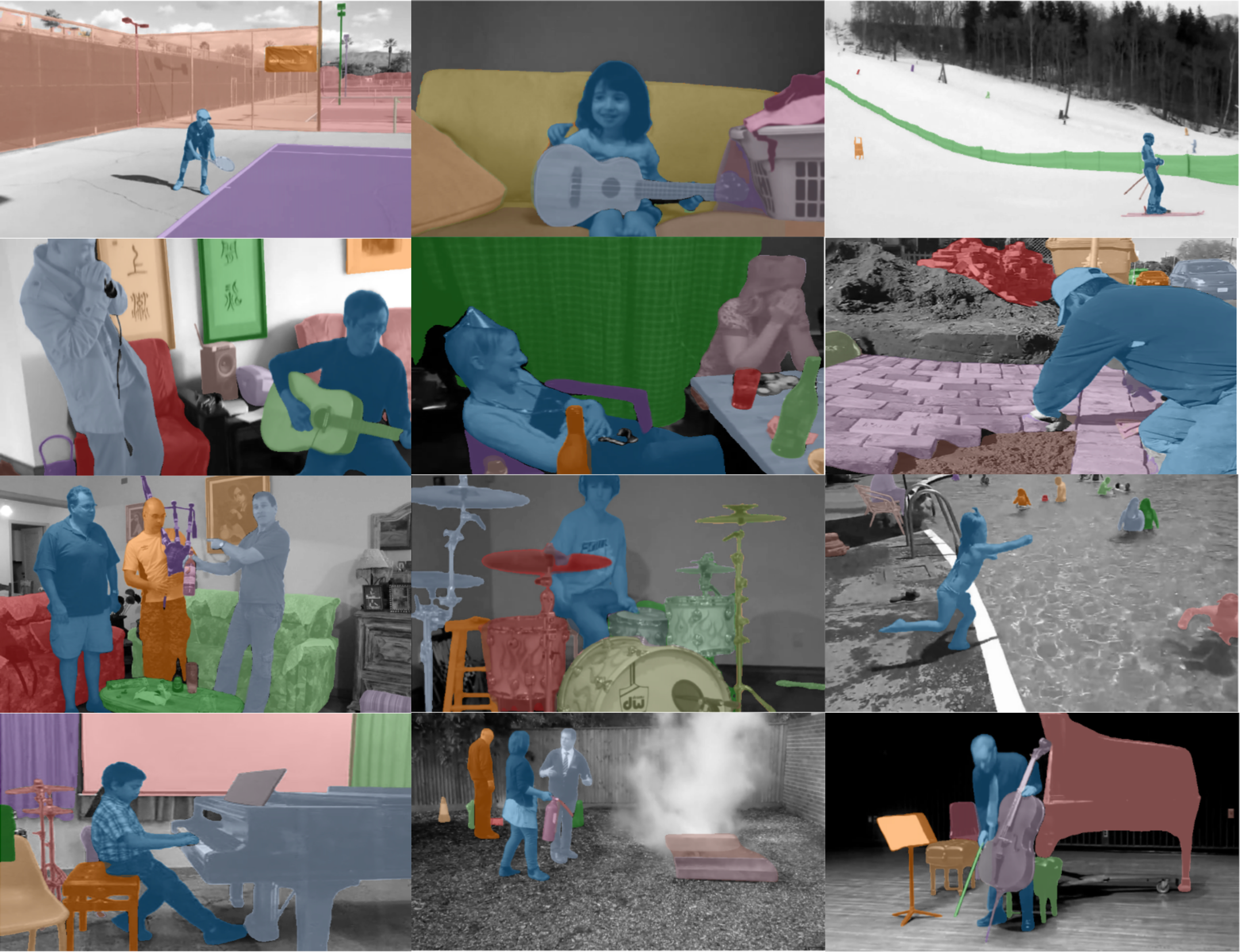}
\caption{UVO testset visualization}
\label{fig:vis}
\end{figure}

We visualize instance segmentation predictions of our method on UVO testset in Fig. \ref{fig:vis}. UVO \cite{uvo} provides 12.29 object annotations per video on average, so we visualize the top 12 confident instances for each image visualization. Note that, for better visual effect, random colors represent different instances rather than categories, UVO is class-agnostic.

\clearpage
\bibliographystyle{splncs04}
\bibliography{egbib}
\end{document}